\documentclass[runningheads]{llncs}
\usepackage{graphicx}
\usepackage{fancyvrb}
\usepackage[ruled,vlined]{algorithm2e}
\usepackage[T1]{fontenc}

\newenvironment{tabularverbatim}
 {\VerbatimEnvironment
  \begin{BVerbatim}[baseline=c,formatcom=\setlength{\baselineskip}{\normalbaselineskip}]}
 {\end{BVerbatim}}

\begin{document}
\title{\textsc{BigCQ}: A large-scale synthetic dataset of competency question patterns formalized into SPARQL-OWL query templates}

\titlerunning{\textsc{BigCQ}: CQ patterns to SPARQL-OWL templates}

\author{Dawid Wiśniewski\inst{1}\orcidID{0000-0003-1194-7921	} \and
Jędrzej Potoniec\inst{1}\orcidID{0000-0002-6115-6485} \and
Agnieszka Ławrynowicz\inst{1}\orcidID{	0000-0002-2442-345X}}
\authorrunning{D. Wiśniewski et al.}
\institute{Faculty of Computing and Telecommunication, Poznan University of Technology\\
\email{\{firstname\}.\{lastname\}@cs.put.poznan.pl}}
\maketitle              % typeset the header of the contribution
\begin{abstract}
Competency Questions (CQs) are used in many ontology engineering methodologies to collect requirements and track completeness and correctness of an ontology being constructed.
Although they are frequently suggested by ontology engineering methodologies, the publicly available datasets of CQs and their formalizations in ontology query languages are very scarce. Since first efforts to automate processes utilizing CQs are being made, it is of high importance to provide large and diverse datasets to fuel these solutions.
In this paper, we present \textsc{BigCQ}, the biggest dataset of CQ templates with their formalizations into SPARQL-OWL query templates. \textsc{BigCQ} is created automatically from a dataset of frequently used axiom shapes. These pairs of CQ templates and query templates can be then materialized as actual CQs and SPARQL-OWL queries if filled with resource labels and IRIs from a given ontology.
We describe the dataset in details, provide a description of the process leading to creation of the dataset and analyze how well the dataset covers real-world examples. We also publish the dataset as well as scripts transforming axiom shapes into pairs of CQ patterns and SPARQL-OWL templates, to make engineers able to adapt the process to their particular needs.

%Query templates can 

%These pairs of pattern with template define domain-agnostic SPARQL-OWL query shapes which target frequently reused OWL constructions and provide general surface forms of questions matching these constructions. 
%These can be then materialized into actual questions and queries if an existing ontology is provided. 

\end{abstract}

\section{Introduction}

Ontologies are formal representations of a given domain of interest used to model entities and relationships between them. They are often expressed using logic-based representations, out of which Web Ontology Language(OWL)~\cite{antoniou2004web} is the most commonly used. The incorporation of logic-based formalization allows to define knowledge precisely and to infer new facts, even if they are not explicitly modelled. Unfortunately, engineers struggle to use such formalization, because it requires proficiency in formal logic. Especially, the logical consequences of modelled knowledge are hard to predict. Also, describing a large, vocabulary-rich domain of interests requires to control how much of vocabulary is already modelled. To address these issues, multiple ontology engineering methodologies guiding engineers how to model knowledge are proposed. They differ in the usage context, but they all share the same goal of making ontology development easier. 

Many ontology engineering methodologies adopted the idea of defining a set of natural language questions, called Competency Questions (CQs), that a finished ontology should be able to answer to correctly. These CQs can be e.g. used to define, as proposed in the NeOn~\cite{Suarez-Figueroa2012} methodology, so-called Glossary-of-Terms defining what vocabulary should be modelled in an ontology and to track the maturity of the ontology by controlling how many CQs can be correctly answered at a given stage of ontology development. If an ontology can answer all CQs it may be regarded as correct and complete.

We have shown~\cite{10.1007/978-3-030-32327-1_36}~\cite{10.1145/3184558.3186575} that an automation of tasks involving CQs, such as automatic Glossary-of-Terms extraction or automatic translation of CQs into an ontology query language targeting T-Box level of the ontology is achievable. 
%Glossary-of-Terms extraction and automation of translation of CQs into ontology query language that targets T-Box level of the ontology is achievable. 
SPARQL-OWL~\cite{DBLP:conf/esws/KolliaGH11} used for querying T-Box level knowledge is a variant of SPARQL~\cite{Harris:13:SQL} with an OWL 2 DL entailment regime. Both problems can be addressed using machine learning based approaches, but these require a huge amount of data to fuel the models with. % Although some datasets exist, they are too small to utilize recent approaches to translation (Seq2Seq neural models) or extraction. 
Although some datasets exists, they are still quite small.
To fill this gap, we introduce \textsc{BigCQ}, a big dataset of CQ patterns mapped to SPARQL-OWL query templates generated automatically from frequently observed axiom shapes extracted from ontologies. These templates can be filled with vocabulary from an ontology to generate a huge number of examples of CQ to SPARQL-OWL query translations in an automatic manner. The CQs alone can be used to create Glossary-of-Terms extractors, while the pairs of CQs and SPARQL-OWL queries can be used to provide a method of automatic translation of CQs into queries.

%In this work we address the following research questions:

%\begin{enumerate}
%    \item \textbf{RQ1}: Is it possible to generate CQs and their SPARQL-OWL formalizations out of frequent axiom %shapes?
%    \item \textbf{RQ2}: What is the coverage of generated CQ patterns measured on the existing CQ sets?
%    \item \textbf{RQ3}: What is the coverage of generated SPARQL-OWL templates measured on the existing SPARQL-OWL %query set?
%    %\item Can \textsc{BigCQ} help to improve some ML-using methods utilizing CQs?
%\end{enumerate}

The dataset that we introduce, as well as all scripts used to create \textsc{BigCQ} are published on GitHub~\footnote{\url{https://github.com/dwisniewski/BigCQ}} and can be cited using DOI~\cite{dwisniewski_2021}. % as well as the dataset of CQ template to SPARQL-OWL query template pairs and a code that may be used to materialize these templates against a given ontology to create pairs of CQs and SPARQL-OWL queries.

%The multitude of concepts and relations to be modeled and the logical consequences of defined knowledge enforce ontology authors to search for methods which can help to measure the quality and matIRIty of ontologies. Many ontology engineering methodologies utilize so-called competency questions to track correctness and completeness of ontologies, but the number of published CQs is rather scarce. The limited number of examples hinders the application of popular large-scale statistical approaches to automate processes involving CQs. The aim of this work is to create a large-scale CQs dataset as well as to provide their formalizations in ontology query language. The dataset is created semi-automatically being created on top of the collection of frequently used axiom shapes, ontology verbalizer (ACE) and template based transformations, which result in a huge mapping of XXX unique CQ shapes mapped to over XXX unique SPARQL-OWL shapes. These, in presence of existing ontologies can be transformed into milions of CQ-query pairs. We hope that such approach will break the data bottleneck of CQ to SPARQL-OWL queries that will result in the adaptation of modern machine learning approaches to automate ontology quality checking.

The remainder of this paper is structured as follows: Section~\ref{sec:relatedwork} describes related work, Section~\ref{sec:existing} describes datasets that were involved in the process of creation and evaluation of the quality of \textsc{BigCQ}. Section~\ref{sec:bigcq} provides details on how the dataset is created. Section~\ref{sec:analysis} provides an analysis of the dataset: its size, features, coverage on existing datasets and discusses cases which are not covered by \textsc{BigCQ}. Section~\ref{impact} discusses the impact of the dataset. Finally, Section~\ref{sec:conclusions} concludes the paper. 
\section{Related work}
\label{sec:relatedwork}
\paragraph{Ontology methodologies utilizing CQs} 
The work of Gruninger and Fox~\cite{Gruninger1995MethodologyFT} was the first methodology for design and evaluation of ontologies that incorporates CQs and use them to define informal questions extracted from ontology motivating scenarios. This idea was later adopted to other methodologies, each of which defining good practices fitting various ontology creation scenarios. Lopez et al. proposed METHONTOLOGY~\cite{FernndezLpez1997METHONTOLOGYFO}, a methodology focusing on an evolving prototype scenario, which, in contrast to classic waterfall-like approaches, allows engineers to move back to previous ontology construction phases. METHONTOLOGY suggest to use CQs to specify requirements for the ontology being built. Suarez-Figueroa et al. presented NeON~\cite{Suarez-Figueroa2012} methodology defining 9 scenarios of ontology creation process. These scenarios address use cases like: creating ontologies from scratch, reusing resources, restructuring or localizing ontological resources. In the context of NeON, CQs are also stated to collect requirements as part of ontology specification phase.
 %The NeON methodology proposes to state CQs to collect requirements as part of ontology specification step. % TDDOnto was proposed by Lawrynowicz et al.~\cite{conf/dlog/LawrynowiczK16}. It focuses on creating tests before modelling knowledge
 
\paragraph{Ontology requirements and their analysis} In recent years several attempts to collect and analyze ontology requirements, defined as natural language CQs and statements have been made. Ren et al.~\cite{DBLP:conf/esws/RenPMPDS14} provided an analysis of the structure of CQs and extracted 19 archetypes of CQs that are defined in terms of CQ templates to be filled with vocabulary (e.g. ‘‘Which [CE1] [OPE] [CE2]?’’). Wisniewski et al.~\cite{DBLP:journals/ws/WisniewskiPLK19} provided an analysis of even bigger dataset of 234 CQs defined for 5 ontologies. Additionally, they provided formalizations of those CQs in terms of SPARQL-OWL queries and analyzed how CQ patterns relate to SPARQL-OWL query templates. The dataset of CQs and their translations was published online. Fernandez-Izquierdo et al.~\cite{10.1007/978-3-030-21348-0_29} published CORAL - a large ontological requirements corpus annotated with lexico-syntactic patterns. This collection of requirements covers both requirements expressed as CQs as well as declarative sentences. % Keet et al.~\cite{DBLP:conf/mtsr/KeetMA19} proposed a controlled natural language that can be used to guide authors on how to state CQs.

\paragraph{Analysis of modelling styles among ontologies}
Lawrynowicz et al.~\cite{DBLP:journals/semweb/LawrynowiczPRT18} provide a way of extracting modeling patterns that recur in ontologies. These frequent formalizations of knowledge are then used to identify emerging ontology design patterns (ODPs), which are solutions to ontology modelling problems. Wang et al.~\cite{DBLP:conf/semweb/WangPH06} analyzed how often OWL language constructs occur in a set of ontologies. Mortensen et al.~\cite{DBLP:conf/semweb/MortensenHMN12} analyzed how ODPs are used in ontologies collected from BioPortal. The authors encoded 68 ODPs using the Ontology PreProcessor Language (OPPL).

\paragraph{Automating processes involving CQs} CQs are used in order to collect requirements and to measure the maturity of the ontology. %Keet et al.~\cite{Keet2019CLaROAC} created a controlled natural language which can be used to guide the construction of CQs to
To automate the process of vocabulary extraction from CQs, Wisniewski et al.~\cite{10.1007/978-3-030-32327-1_36} proposed a machine learning-based tagger, which automatically detects candidates for classes and properties to be modelled in an ontology. Wisniewski also proposed the idea of automatic translation of CQs into SPARQL-OWL queries~\cite{10.1145/3184558.3186575}, so that having an ontology under development, a set of CQs can be used to evaluate the quality of the ontology without the need of manual translation of CQs into query formalization.

%CLaRO and us

%Wisniewski et al.~\cite{10.1007/978-3-030-32327-1_36} proposed an automatic extraction of Glossary of Terms out of CQs.

\paragraph{Question generation} Generating questions automatically from a given portion of text is an NLP task named question generation. The methods addressing this task often use rule-based systems and neural networks to generate questions. Wang et al.~\cite{DBLP:conf/lats/WangLNWGB18} proposed QG-Net, a recurrent neural network-based approach generating quiz questions for educational purposes. Aldabe et al.~\cite{DBLP:conf/its/AldabeLMMU06} used NLP tools and specific linguistic information to generate questions. Liu et al.~\cite{DBLP:conf/its/LiuCR10} proposed an automatic question generator for literature review writing support. This generator is based on a set of templates and the content elements.

\paragraph{Ontology verbalization} Ontologies expressed using OWL are hard to be read by people. To understand the content of ontologies easier, there are several approaches to translate the OWL formalization into natural language proposed, so that axioms from the ontology can be expressed as natural language statements. The most popular approach, ACE verbalizer was proposed by~\cite{kaljurand2007attempto}. Here, axioms are transformed into Attempto Controlled English~\cite{DBLP:conf/rweb/FuchsKK08}, a controlled language reducing complexity of sentences and providing a simple vocabulary to build sentences. The ACE verbalizer, to work properly, assumes that ontolgoy classes and instances have nouns as labels, and properties are named using verbs.

%\paragraph{Natural language processing - dependency trees}

\section{Existing datasets used to produce and evaluate \textsc{BigCQ}}
\label{sec:existing}

%In this section, we describe existing datasets that influenced the \textsc{BigCQ} creation process.

\subsection{Frequent axiom patterns}
The dataset of frequent axiom patterns introduced by Lawrynowicz et al.~\cite{DBLP:journals/semweb/LawrynowiczPRT18} and published online~\footnote{\url{https://semantic.cs.put.poznan.pl/bioportal-patterns/}} is the main inspiration of \textsc{BigCQ}. To fit our needs, we preprocessed each axiom pattern in the dataset, replacing all missing fragments and all vocabulary (apart of RDF, RDFS, OWL and XSD) with IRIs coming from \texttt{http://example.ns} namespace. These IRIs encode, as part of local names, the resource type and numerical identifier allowing to detect when the same IRI is referenced multiple times in one axiom. All kinds of artificial IRIs are defined in Table~\ref{placholderdefinitions}. As part of this process, we also serialized each axiom using Turtle language~\footnote{\url{https://www.w3.org/TR/turtle/\#language-features}}.
This procedure transformed frequent axiom patterns mined from BioPortal into domain-agnostic shapes, so that variables and domain-dependent IRIs are removed, leaving only the axiom shape providing information on how are resources frequently related. We expect that these shapes may be often reused when designing new ontologies.

\begin{table}
\begin{center}
\caption{The types of artifical IRIs used in frequent axiom shapes.}
\label{placholderdefinitions}
 \begin{tabular}{|l|l|r|} 
 \hline
 artificial IRI & short name (label) & meaning \\ 
 \hline\hline
 \texttt{http://example.ns\#C\{NUM\}} & [C\{NUM\}] & a class \\ 
 \hline
 \texttt{http://example.ns\#I\{NUM\}} & [I\{NUM\}] & an individual \\
 \hline
 \texttt{http://example.ns\#OP\{NUM\}} & [OP\{NUM\}] & an object property \\
 \hline
 \texttt{http://example.ns\#DP\{NUM\}} & [DP\{NUM\}] & a data property \\
 \hline
 \texttt{http://example.ns\#DT\{NUM\}} & [DT\{NUM\}] & a data type \\
 \hline

\end{tabular}
\end{center}
\end{table}

%\begin{table}
%\begin{center}
% \begin{tabular}{|l|l|l|r|} 
% \hline
% id & Group description & example & count \\ 
% \hline\hline
% 1 & Equiv+AD+AR & Every X is Y and Every Y is X & 0 \\ \hline
% 2 & Equiv+AD+CR & Every X is Y and Every Y is X & 0 \\ \hline
% 3 & Equiv+CD+AR & Every X is Y and Every Y is X & 0 \\ \hline
% 4 & Equiv+CD+CR & Every X is Y and Every Y is X & 0 \\ \hline
% 5 & SubClass+AD+AR & Every X is Y and Every Y is X & 0 \\ \hline
% 6 & SubClass+AD+CR & Every X is Y and Every Y is X & 0 \\ \hline
% 7 & SubClass+CD+AR & Every X is Y and Every Y is X & 0 \\ \hline
% 8 & SubClass+CD+CR & Every X is Y and Every Y is X & 0 \\ \hline
%
%\end{tabular}
%\end{center}
%\caption{Groups of class axioms and the complexity of their domain and range. \textbf{AD} and \textbf{AR} represent atomic (single named class) domain of class axiom and range of class axiom respectively. \textbf{CD} and \textbf{CR} represent class expressions being constructed with more than one resource in domain and range of class axiom respectivley.}
%\label{groups}
%\end{table}

As a result of this procedure, 239 axiom shapes were collected.
%that was constructed from axioms extracted from on a set of 331 ontologies from the BioPortal repository. Each axiom was transformed into so-called axiom shape, by replacing actual IRIs with artificial ones, so that similar constructs can be spotted among different places in an ontology or among different, covering different domains, ontologies. Every artificial IRI sharing common domain and encoding the type of the resource as part of the local name. Every local name ends with numerical suffix, which is useful to identify cases in which the same resource is referenced multiple times in an axiom. Table~\ref{placholderdefinitions} provides a list of IRIs used to represent different resource types. For further convenience, we also define short names (labels) for every artificial IRI.
An example of a frequently reused OWL axiom shape is:

\begin{verbatim}
<http://example.ns#OP1> a owl:ObjectProperty .
<http://example.ns#C1> a owl:Class .

<http://example.ns#C2> a owl:Class; rdfs:subClassOf [
    a owl:Restriction;
    owl:onProperty <http://example.ns#OP1>;
    owl:hasValue <http://example.ns#C1>; ] .
\end{verbatim}

%The \textsc{BigCQ} is generated based on these axiom shapes collection, by transforming each shape into CQ to SPARQL-OWL query template pairs.

%\subsection{Datasets of CQs and their formalizations into SPARQL-OWL}

\subsection{Existing datasets of requirements and their formalizations}
The dataset introduced by Wisniewski et al.~\cite{DBLP:journals/ws/WisniewskiPLK19} (further referenced as \textsc{CQ2SPARQLOWL}) and their analysis on how CQs are constructed and how they are related to provided SPARQL-OWL queries was our main guide on how to relate CQs to queries.
Because in \textsc{BigCQ} SPARQL-OWL query templates are produced from axiom shapes by applying simple substitutions of IRIs with variables followed by preamble and postamble adding and because the ontologies used to generate axiom shapes are disjoint with ontologies in \textsc{CQ2SPARQLOWL}, we used all SPARQL-OWL queries from \textsc{CQ2SPARQLOWL} to measure the coverage of \textsc{BigCQ} query templates.

%translate axioms into queries and CQs. Because SPARQL-OWL query templates are generated by changing resource IRIs with variables and adding appropriate pre- and postambles based on shapes extracted from BioPortal, we can use SPARQL-OWL queries from \textsc{CQ2SPARQL} to evaluate the coverage of generated queries fairly. 
However, \textsc{CQ2SPARQLOWL} was used intensively to observe how existing queries relate to real CQs. Because of that, in order to evaluate the coverage of CQ templates we used another dataset of CQs to check how well our CQ templates cover unseen, real CQs. The \textsc{CORAL} dataset introduced by Fernandez-Izquierdo et al.~\cite{10.1007/978-3-030-21348-0_29} is the biggest compilation of requirements expressed as either CQs or declarative sentences. The dataset overlaps with \textsc{CQ2SPARQLOWL}, so we used only CQs that were not present in \textsc{CQ2SPARQLOWL} as an evaluation set to calculate the coverage of CQ templates from \textsc{BigCQ}.
%From that dataset, we used requirements expressed as CQs, that were not a part of \textsc{CQ2SPARQL}. Since there is an overlap between \textsc{CQ2SPARQLOWL} and \textsc{CORAL}, we selected those being unique for \textsc{CORAL} to test only on previously unseen cases.

% We used the dataset introduced by Wisniewski et al.~\cite{DBLP:journals/ws/WisniewskiPLK19} (further referenced as \textsc{CQ2SPARQLOWL}) and the analysis they provided to get inspiration on how CQs relate to SPARQL-OWL queries. We also use CQs provided in  \textsc{CORAL}, the biggest requirements dataset, which is used as an evaluation dataset, to see how well our constructed CQs cover unseen cases.

%Because \textsc{CORAL} contains requirements expressed as CQs and as declarative statements and, because there is a non-zero overlap between CQs from \textsc{CQ2SPARQLOWL} and \textsc{CORAL}, to validate the quality of our dataset we use only a subset of \textsc{CORAL}: only those requirements that are CQs and that were not part of \textsc{CQ2SPARQLOWL}.

\begin{table}
\begin{center}
\caption{The size of requirement datasets involved in \textsc{BigCQ} construction and evaluation.}
 \begin{tabular}{|l|l|} 
 \hline
 Dataset & Number of CQs\\ 
 \hline\hline
 \textsc{CORAL} (all CQs + sentences) & 834\\
 \hline
 \textsc{CORAL} (all CQs) & 469\\ 
 \hline
 \textsc{CORAL} (all CQs that are not in \textsc{CQ2SPARQLOWL}) & 324\\ 
 \hline
 
 \textsc{CQ2SPARQLOWL} (all CQs)  & 234\\
 \hline
 \textsc{CQ2SPARQLOWL} (all CQs with SPARQL-OWL queries defined) & 131\\ 
 \hline

\end{tabular}
\end{center}
\label{datasetsizes}
\end{table}

%To perform an analysis of existing CQs and to evaluate our created CQ forms we decided to use two available datasets. The dataset by Wisniewski et al. [REFERENCE] contains 234 CQs stated against terminological layer of 5 different ontologies. [TODO] of these CQs are paired with their formalizations in SPARQL-OWL. The analysis provided by the paper accompanying the dataset shows that frequently CQs are formalized similarly among ontologies. 

%The dataset named CORAL was used to evaluate how well our method covers CQs that were previously unseen. Since \textsc{CORAL} contains requirements expressed as both statements and questions, since the scope of the paper are CQs only, we ignored statements. 
%Moreover, a subset of CORAL is also a part of CQ2SPARQLOWL, because of that, we preserved only those CQs from Coral that were not observed ini CQ2SPARQLOWL.

%ACE Verbalizer

\section{\textsc{BigCQ} - a dataset of CQ templates matched with corresponding SPARQL-OWL formalization templates}
\label{sec:bigcq}

In this section we describe the process of transformation of each axiom shape into a pair of CQ template and SPARQL-OWL template. To achieve this goal, we first verbalize each axiom shape to obtain statements which can be easily transformed later into questions. Then each axiom shape is transformed into SPARQL-OWL query templates by adding appropriate preamble and postamble and replacing IRIs with variables.

%In this section we describe the process of transformation of each frequent axiom shape into a textual form, which is a convenient intermedie representation form guiding CQs to be constructed. The pairs of axiom shapes and verbalizations are then analyzed to find common features allowing to identify groups of similarly constructed CQs and queries. Finally, we describe how we create SPARQL-OWL queries out of axiom shapes, how do we transform verbalizations into CQs and how do we relate CQs to corresponding queries.

\subsection{Verbalizing axiom shapes}
\label{verbalizing_intro}
%To understand the dataset better, and to ease the process of stating CQ, we use ACE verbalizer to express each OWL axiom shape as statements in natural language. Out of 239 axiom shapes in total, 
%we obtained 125 unique verbalizations. For 85 axiom shapes, ACE failed to produce a verbalization, returning \texttt{BUG: Axiom too complex} error instead.
%182. For 85 axiom shapes, ACE failed to produce a valid output, returning \texttt{BUG: Axiom too complex} error instead. From all 182 verbalizations 125 was unique. For further analysis and transfromations, we used all 125 verbalizations.

%Each of the 239 axiom shapes was verbalized using ACE verbalizer to generate natural language statements. We introduce this intermediate step to ease the task of formulating CQs -- for human beings it is easier to state questions for sentences rather than formalizations. \textbf{TODO - rewrite that last sentence, link to appendix with technical deails on this step}. As a result of this step, we have collected 125 unique verbalizations and axiom shapes assigned to each verbalization.

Using ACE verbalizer~\footnote{\url{https://github.com/Kaljurand/owl-verbalizer}}, we verbalized each of the 239 axiom shapes into natural language statements. An example OWL axiom shape from the dataset:

\begin{verbatim}
<http://example.ns#OP1> a owl:ObjectProperty .
<http://example.ns#C1> a owl:Class .
<http://example.ns#C2>  a owl:Class ; rdfs:subClassOf [
    a owl:Restriction;
    owl:onProperty <http://example.ns#OP1>;
    owl:hasValue <http://example.ns#C1> ; ] .
\end{verbatim}

is verbalized by ACE verbalizer into \texttt{Every C2 OP1 C1} statement.

\subsection{Analysis of axiom shapes and their formalizations}
\label{axiomanalysis}

%Each axiom shape collected uses either \texttt{rdfs:subClassOf} or \texttt{owl:equivalentClass} class axiom type to build classes from some transformations of other classes. Let us define class axiom domain (CAD) and class axiom range (CAR) as the left-hand and right-hand argument of class axiom respectively.

\begin{figure}[htbp]
    \centering
    \includegraphics[width=\textwidth]{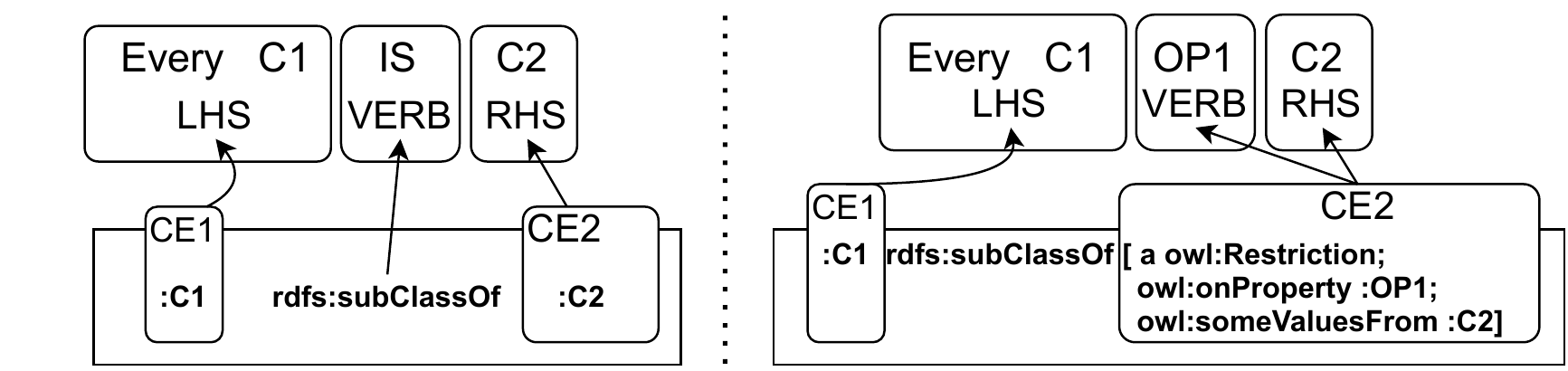}
    \caption{Two examples of simple verbalizations showing how mapping between class expressions(CE1, CE2) and parts of the verbalization (LHS/RHS/VERB) is performed.}
    \label{fig:cece}
\end{figure}

Each axiom shape collected uses either \texttt{rdfs:subClassOf} or \texttt{owl:equivalentClass} to relate two class expressions, which can be simple named classes or complex class expressions. In general any axiom collected can be expressed as \texttt{CE1 rdfs:subClassOf CE2} or \texttt{CE1 owl:equivalentClass CE2}, where \texttt{CE\{NUM\}} represents a given class expression. Each verbalization of collected axiom shapes follow the  \{LHS\} \{VERB\} \{RHS\} pattern, where \{VERB\} is a main verb (the root of the dependency tree constructed from the verbalization\footnote{In order to to build valid dependency trees from verbalizations, we temporality (only for the task of dependency tree building) substituted artificial property names (op1, dp1, \ldots) with verbs, so that correct part of speech tags can be assigned to tokens and correct dependency trees can be constructed.}), and \{LHS\} (left-hand-side) and \{RHS\} (right-hand-side) are related to class expressions \texttt{CE1} and \texttt{CE2} from the axiom shape that was verbalized. If \texttt{CE2} begins with a property restriction (as can be observed in an example in Section~\ref{verbalizing_intro}), the \{VERB\} becomes the property label (which ACE assumes to be some verb\footnote{\url{http://attempto.ifi.uzh.ch/site/docs/owl\_to\_ace.html}}), otherwise, the \{VERB\} becomes a word \texttt{is} representing subsumption relation between two class expressions. These two possible mappings are visualized in Figure~\ref{fig:cece}.

The distinction between what \{VERB\} can be defines two classes of verbalizations:
\begin{enumerate}
    \item expressing subject-property-object (SPO) relation: All verbalizations linking left-hand-side (LHS) and right-hand-side (RHS) of the main verb (VERB) with a property label express some non-taxonomical relation (e.g. a verbalization of \texttt{Every C1 OP1 C2}, further materialized with ontology vocabulary into e.g. \texttt{Every computer executes code} states that \texttt{computer} has the ability to \texttt{execute} \texttt{code}).
    %\item subject being linked to object by a property (subject-property-object (SPO) for short) (e.g. \texttt{Every C1 OP1 C2}, which can be materialized further into e.g. \texttt{Every computer executes code}).
    \item expressing subclass-superclass (SS) relation: If LHS and RHS of the main verb are linked with \texttt{is} verb, the verbalization expresses a subsumption relation between two class descriptions (e.g. \texttt{Every C1 is C2}, further materialized into e.g. \texttt{Every programmer is a person}).
\end{enumerate}

\subsubsection{Transformation from SPO to SS}

It is worth to mention that the SPO case can be transformed into the SS by injecting \texttt{is something that} before the main verb (property label). \texttt{Every C1 OP1 C2 (Every computer executes code)} can be transformed into \texttt{Every computer is something that executes code}. Using this transformation \texttt{is} becomes the new main verb (the new dependency-tree root) and \texttt{something that executes code} as a whole becomes RHS of this verb. %This paraphrase is in line with axiom shape the verbalization was created from. Even if the initial verbalization used property label as the main verb, the corresponding axiom shape used \texttt{rdfs:subClass} to relate its CAD to its CAR. \textbf{TO MOZE BYC NIEJASNE DLA CZYTAJACEGO}

%\subsubsection{Relation between CAD, CAR and verbalization left and right hand sides }

%The relation between CAD and CAR and between main verb's (dependency tree root) LHR and RHS is defined as follows:
%\begin{itemize}
%    \item verbalized CAD equals main verb's LHS.
%    \item verbalized CAR equals:
%    \begin{enumerate}
%        \item main verb's RHS if the main verb is represented by \texttt{is} word.
%        \item main verb's RHS plus the main verb itself. if the main verb is a property name used instead of \texttt{is}.
%    \end{enumerate}
%\end{itemize}

%Figure~\ref{fig:carcad} presents an example of a frequent axiom shape, its verbalizations and relations between CAD, CAR and left-hand-side and right-hand-side of the main verb.

\subsubsection{\textsc{BigCQ} design choices}

\begin{figure}[h]
    \centering
    \includegraphics[width=0.7\textwidth]{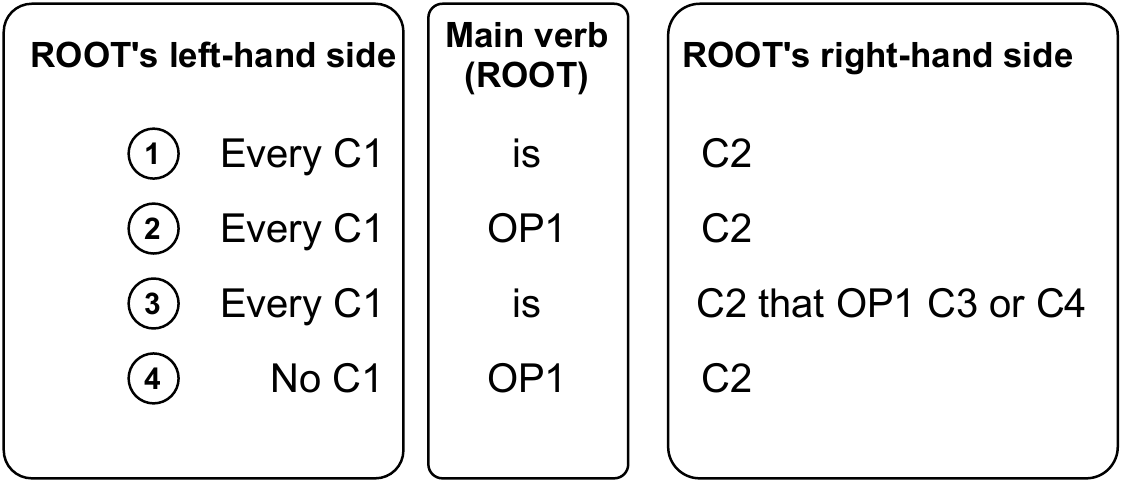}
    \caption{4 examples of verbalizations created from axiom shapes, split into \{LHS\}, \{VERB\} (the root node of the dependency tree) and \{RHS\} parts.}
    \label{fig:CARCAD}
\end{figure}

%Having defined all basic concepts and relation between parts of axiom shapes and verbalizations, we can analyze the dataset and make the following observations:
An in-depth analysis of axiom shapes and their verbalizations revealed the following characteristics of the data collected:
\begin{itemize}
    \item There are 2 main groups of axiom shapes, 73 of them use \texttt{owl:equivalentClass} as the class axiom type, while others use \texttt{rdfs:subClassOf}. Stating CQs regarding the first group, we should include words and phrases like: identical, the same, equal, etc. On the other hand, stating CQs regarding the second group, we should include words and phrases like: a kind of, a type of, a specialization of, etc.
    \item The distinction between SPO and SS classes of verbalizations is required to state grammatically correct CQs. Forms of CQs that can be applied to cases following the SPO relation, may not work well for the SS. For instance, a sentence: \texttt{Every animal eats grass} (an instantiation of \texttt{Every C1 OP1 C2}) can be transformed into a question: \texttt{Does every animal eat grass?}. However, the same transformation cannot be applied to a sentence: \texttt{Every animal is a living being} (an instantiation of \texttt{Every C1 is C2}). We can rephrase the SPO to the SS by injecting \texttt{is something that} and then use procedure defining transformations of statements into questions that were defined for the SS case. %to handle such transformed SPO verbalizations.
    \item Although we can construct questions and queries that target any combination of resource labels and resource IRIs, we decided to simplify the problem:
    \begin{enumerate}
        \item Among the \textsc{CQ2SPARQLOWL} dataset, less than 2\% of CQs ask for more than one resource at once. Thus, we decided that all CQ templates and query templates in \textsc{BigCQ} are stated with at most one resource as a query target.
        \item In general we can ask for any resource stated in an axiom, but for axioms involving complex class expressions it becomes hard to produce a question regarding resources in these complex class expressions. Figure~\ref{fig:CARCAD} presents 4 examples of verbalizations. First two of them use named classes in both LHS and RHS. In these cases it is trivial to state questions targeting resources C1 and C2. In example 1, we can state e.g. \texttt{What are the kinds of C2?} to taget resource C1 or state \texttt{What is the name of the category that C1 falls into?} to target resource C2. Similarly, we can state \texttt{What things OP1 C2?} to target resource C1 in example 2 and \texttt{What things does C1 OP1?} to target resource C2. But if a resource is a part of a complex class expression, like resource C1 in example 4 or resources C2, C3 or C4 in example 3, it becomes very hard to produce a question targeting these resources. Even if if is possible, such questions are likely to be very complicated. Considering above, we decided that we will create CQs and queries targeting the whole (i) LHS if it contains a single named class, (ii) RHS if it contains a single named class (iii) VERB if it is a property label (in that case, we can state questions like \texttt{What relates C1 and C2?}).
        
        %Although it is very easy to state a question about LHS and RHS if the is only one named class involved (e.g. regarding example 1 from that figure, \texttt{What are the kinds of C2?} and \texttt{What is the name of the category that C1 falls into?}) can be used to query LHS and RHS respectively, if LHS or RHS becomes a description involving more than one resource, the questions to be stated become more complicated. In example 4, it is not clear how to construct a CQ querying LHS (No C1). In example 3, considering particular resources in its RHS, it is not obvious how to state a question targeting C2, OP1, C3 or C4. Even if this is technically possible, the CQs generated will become very unnatural. Considering above, we decided to generate questions and queries targeting (LHS, RHS) only if it (LHS, RHS) consists of a single named class. Moreover, we can create questions about verbs if those are properties. In examples from Figure~\ref{fig:CARCAD}, \textsc{BIGCQ} can ask about: example 1: LHS, RHS, example 2: LHS, VERB, RHS, example 3: LHS, example 4: VERB, RHS. Even if the simplification lowers the number of CQ templates that can be generated, we can still obtain many templates. We can ask about LHS of example 1 from Figure~\ref{fig:CARCAD} with \texttt{What is the specialization of C2?} and ask about example 3 with \texttt{What is the specialization of C2 that OP1 C3 or C4}.
    \end{enumerate}
    
   % There are two possible of main verbs used in verbalizations. If a verbalization follows SPO group, the questions stated should be different than in SS group. An example \texttt{Every C1 OP1 C2} can be translated into \texttt{Does every C1 OP1 C2?}. An example \texttt{Every C1 is C2} can be translated into \texttt{Is every C1 C2?}. Using transformation of SPO into SS, we can express \texttt{Every C1 OP1 C2} verbalization as \texttt{Is every C1 something that OP1 C2?}.
    %\item Binary questions regarding SS type can be expressed using e.g. \texttt{Is [S] [O]?}, where \texttt{S} and \texttt{O} are subject and object respectively.
    %\item Binary questions regarding SPO type can be expressed using \texttt{Is it true that [S] [P] [O]?}, where \texttt{S}, \texttt{P} and \texttt{O} represent subject, predicate and object respectively. Alternatively, we can transform it into \texttt{Is it true that [S] is something that [P] [O]}.
    %\item axiom shapes using \texttt{owl:equivalentClass} class axiom type, when translated to CQs, 
\end{itemize}

\subsection{SPARQL-OWL query shapes generation}
\label{queryshapes}
The SPARQL language standard\footnote{\url{https://www.w3.org/TR/rdf-sparql-query/\#QueryForms}} defines 4 forms of SPARQL queries that can be stated: 
\begin{itemize}
    \item ASK - used to check if a given query pattern is matched in the ontology.
    \item SELECT - used to select the IRIs of variables bound in a query pattern match.
    \item DESCRIBE - used to obtain RDF graphs that describe the resources found.
    \item CONSTRUCT - used to obtain RDF graphs constructed by substituting variables in a set of triple templates.
\end{itemize}

\textsc{CQ2SPARQLOWL}, the biggest to date dataset of CQ to SPARQL-OWL translations utilizes only two forms of queries: \texttt{ASK} and \texttt{SELECT}. Both types return results that are easy to be interpreted by humans, since \texttt{ASK} queries return a boolean answer and matches binary (yes/no) questions, while SELECT queries return resource identifiers and matches Wh- questions. In contrast, \texttt{DESCRIBE} and \texttt{CONSTRUCT} return RDF-graphs that show how the knowledge is modelled rather than what is modelled and we consider them out of scope for the CQs usage. In this section, we describe how axiom shapes are transformed into \texttt{ASK} and \texttt{SELECT} SPARQL-OWL query shapes.

Before transforming each axiom shape into SPARQL-OWL query, from each axiom shape we removed declarations associating types to IRIs and replaced IRIs with short placeholders encoding resource local name enclosed with <>. These placeholders contain identifiers required to link them with CQ templates.

There are 7 types of queries that can be stated if we follow the simplification mentioned at the end of Section~\ref{axiomanalysis}. All types, as well as examples of SPARQL-OWL query shapes are listed in Table~\ref{tab:types}. As can be seen, ASK query is produced simply by wrapping the axiom shape with \texttt{ASK WHERE \{ ... \}}, SELECT queries by wrapping the axiom shape with \texttt{SELECT ?x WHERE \{ ... \}} and transforming resources (if they are single named classes) related to \{LHS\}, \{RHS\} and \{VERB\} into a variable named \texttt{?x} and SELECT COUNT queries are produced  out of SELECT queries by further substituting \texttt{SELECT ?x} with \texttt{SELECT COUNT (?x)}.

Each axiom shape can be transformed to at most 7 queries. This is possible if both \{LHS\} and \{RHS\} of the verbalization describe single named classes and if \{VERB\} is a property label. If \{LHS\} and \{RHS\} are both complex expressions, and the main verb is represented by \texttt{is}, then, only ASK query is produced.

\begin{table}[htbp]
 \small
\begin{center}
\caption{All supported forms of questions generated from \texttt{<C1> rdfs:subClassOf [a owl:Restriction; owl:onProperty <OP1>; owl:someValuesFrom <C2>]} axiom shape.}
\label{tab:types}
 \begin{tabular}{|p{2cm}|p{12cm}|}
 \hline
Question Type & Query shape example \\ 
 \hline
ASK & 
\begin{tabularverbatim}

ASK WHERE {<C1> rdfs:subClassOf [a owl:Restriction;
                                 owl:onProperty <OP1>;
                                 owl:someValuesFrom <C2>]}
\end{tabularverbatim} 

\\
\hline
SELECT LHS & 
\begin{tabularverbatim} 

SELECT ?x WHERE {?x rdfs:subClassOf [a owl:Restriction; 
                                     owl:onProperty <OP1>;
                                     owl:someValuesFrom <C2>]} 
\end{tabularverbatim}

\\
\hline
SELECT RHS & 
\begin{tabularverbatim}

SELECT ?x WHERE {<C1> rdfs:subClassOf [a owl:Restriction;
                                       owl:onProperty <OP1>;
                                       owl:someValuesFrom ?x]} 
\end{tabularverbatim}

\\
\hline
SELECT VERB & 
\begin{tabularverbatim}

SELECT ?x WHERE {<C1> rdfs:subClassOf [a owl:Restriction;
                                       owl:onProperty ?x;
                                       owl:someValuesFrom <C2>]} 
\end{tabularverbatim}
\\
\hline
SELECT COUNT LHS &
\begin{tabularverbatim}

SELECT COUNT(?x) WHERE {?x rdfs:subClassOf [a owl:Restriction; 
                                            owl:onProperty <OP1>;
                                            owl:someValuesFrom <C2>]} 
\end{tabularverbatim}

\\
\hline
SELECT COUNT RHS &
\begin{tabularverbatim}

SELECT COUNT(?x) WHERE {<C1> rdfs:subClassOf [a owl:Restriction;
                                              owl:onProperty <OP1>;
                                              owl:someValuesFrom ?x]}
\end{tabularverbatim}

\\
\hline
SELECT COUNT VERB &
\begin{tabularverbatim}

SELECT COUNT(?x) WHERE {<C1> rdfs:subClassOf [a owl:Restriction;
                                              owl:onProperty ?x;
                                              owl:someValuesFrom <C2>]} 
\end{tabularverbatim}

\\
\hline
\end{tabular}
\end{center}
\end{table}

\subsection{Statements to CQ templates transformation}
For each query type defined, we handcrafted a set of transformations that can be used to transform a given verbalization (statement) into CQ templates depending on a given question type. In Table~\ref{tab:cqtemplates}, we provide a single example for each category for verbalizations following the SPO type, if a corresponding axiom shape does not use \texttt{owl:equivalentClass}.
%\begin{itemize}
%    \item \texttt{ASK} -> \texttt{Does {LHS} {VERB} {RHS}}?
%    \item \texttt{QUERY TARGET LHS} -> \texttt{What {VERB} {CAR}?}
%    \item \texttt{QUERY TARGET RHS} -> \texttt{What {CAD} {VERB}}?
%    \item \texttt{QUERY target VERB} -> \texttt{What relates {CAD} and {CAR}?}
%    \item \texttt{COUNT QUERY TARGET LHS} -> \texttt{How many things {VERB} {CAR}}?
%    \item \texttt{COUNT QUERY TARGET RHS} -> \texttt{How many things {CAD} {VERB}}?
%    \item \texttt{COUNT QUERY TARGET VERB} -> \texttt{How many relations are there between {CAD} and {CAR}?}
%\end{itemize}

\begin{table}[htbp]
\begin{center}
\caption{Examples of transformations applicable for verbalizations, among which \{VERB\} is a property label} %and % axiom shape does not define class equivalence.}
\label{tab:cqtemplates}
 \begin{tabular}{|l|p{7cm}|} 
 \hline
 Question type & Example of CQ template \\ 
 \hline
 \texttt{ASK} & \texttt{Does \{LHS\} \{VERB\} \{RHS\}?} \\ \hline
 \texttt{QUESTION TARGET LHS} & \texttt{What \{VERB\} \{RHS\}?} \\ \hline
 \texttt{QUESTION TARGET RHS} & \texttt{What does \{LHS\} \{VERB\} ?} \\ \hline
 \texttt{QUESTION TARGET VERB} & \texttt{What relates \{LHS\} and \{RHS\}?}\\ \hline
 \texttt{QUESTION TARGET COUNT LHS} & \texttt{How many things \{VERB\} \{RHS\}?} \\ \hline
 \texttt{QUESTION TARGET COUNT RHS} & \texttt{How many things does \{LHS\} \{VERB\}?} \\ \hline
 \texttt{QUESTION TARGET COUNT VERB} &\texttt{How many relations are there between \{LHS\} and \{RHS\}?} \\ 
 \hline
\end{tabular}
\end{center}
\end{table}

The \{LHS\}, \{VERB\}, \{RHS\} markers introduced among among these examples are replaced with fragments of verbalizations that were marked as \{LHS\}, \{VERB\}, \{RHS\}. Using such a replacement, we can, even if a single transformation type provided, generate multiple CQ templates if verbalizations provide various \{LHS\}, \{VERB\}, \{RHS\} forms. 

%Using such approach we can obtain multiple CQ templates even using a single CQ template (e.g. Having two verbalizations \texttt{Every C1 is C2} and \texttt{Every C1 is C2 that OP1 C3}, their \{RHS\} will be different so different sets of CQ templates will be produced).

%. % If different fragments of texts will represent placeholders, they will create different CQ templates.

%The placeholders \texttt{{LHS}}, \texttt{{VERB}}, \texttt{{RHS}} are substituted with left-hand-side, main verb and right-hand-side extracted from a verbalization of a given axiom shape to generate question templates. Thus, for axiom shape \texttt{Every C1 OP1 C2}, the ask query will be \texttt{Does C1 OP1 C2?} and if the axiom shape is \texttt{Every C1 OP1 C2 that OP3 nothing but C4}, then the ask query will be of form \texttt{Does C1 OP1 C2 that OP3 nothing but C4}. If RHS is a complex class description, we do not generate \texttt{QUERY TARGET RHS} and \texttt{COUNT QUERY TARGET RHS}, since the answer is not a single class. Similarly if CAD is a complex class description, we do not generate \texttt{QUERY TARGET LHS} and \texttt{COUNT QUERY TARGET LHS}.

We defined separate transformations for verbalizations expressing equivalence (where questions should ask for equiality between classes) and for those expressing subsumption. We also defined separate transformations for verbalizations following the SPO and the SS classes.

Similarly to the process defined in Section~\ref{queryshapes}, if both \texttt{\{LHS\}} and \texttt{\{RHS\}} represent single named classes and the main verb is a property label, we construct lists of CQs for each of 7 possible question types. However, if \texttt{\{LHS\}} or \texttt{\{RHS\}} defines a complex class expression or if the main verb is simply \texttt{is}, only a subset of the 7 question types will be addressed. In the worst case scenario (both \texttt{\{LHS\}} and \texttt{\{RHS\}} represent complex class expression and the main verb is \texttt{is}, we state only ASK questions.

\subsection{Synonyms substitution in CQ templates}
In English, there are words or phrases that can be used interchangeably. For example \texttt{a type}, \texttt{a kind}, \texttt{a specialization} all share the same meaning. We use the interchangeability of words to introduce synonym sets into our CQ templates. Transformations used to produce CQ templates can introduce special markers, which if found, are replaced with all possible synonyms in the synonym set.

For example, a CQ template: \texttt{[What] C1 are [types] of C2?} contains two synonym set markers [What] and [types]. These tokens are mapped to predefined synonyms lists, eg: [What] = What or which, [types] = types or kinds. If synonyms substitution is applied, these markers are replaced with all possible forms to generate multiple CQ templates: (i) \texttt{What C1 are types of C2?} (ii) \texttt{Which C1 are types of C2?} (iii) \texttt{What C1 are kinds of C1?} (iv) \texttt{Which C1 are kinds of C2?}. The more synonym set markers occur and the more synonyms they list, the more paraphrases of questions we can generate.

\subsection{Relating CQ templates to SPARQL-OWL query templates}
Every verbalized axiom shape is transformed using predefined transformations into CQ templates and then paraphrases are generated during synonym substitution phase. These CQ templates are linked during transformation phase to 7 categories introduced in Table~\ref{tab:cqtemplates}. Similarly, SPARQL-OWL query templates from axiom shapes are also linked to 7 related categories introduced in Table~\ref{tab:types}. We pair all CQ templates with query templates sharing the same category.

\section{Dataset analysis}
\label{sec:analysis}

The \textsc{BigCQ} dataset is summarized in Table~\ref{tab:datasetsummary}. As can be seen, \textsc{BigCQ} provides over 77575 distinct CQ templates. This is over 93x more than the size of the biggest to date requirements dataset name \textsc{CORAL}. However, CQ templates should be filled with vocabulary from actual ontologies to create CQs, so regarding all possible ways of filling templates based on vocabulary from an ontology, we can generate a dataset of CQs being thousands or millions times bigger than \textsc{CORAL}. Also, regarding the number of SPARQL-OWL query templates, 549 query templates provided in \textsc{BigCQ} is much more than the number of SPARQL-OWL queries defined in the biggest dataset of SPARQL-OWL queries \textsc{CQ2SPARQL} (\textsc{CQ2SPARQLOWL} defines 131 queries, out of which many share the same SPARQL-OWL query template). Also here, there are multiple ways of filling templates with resource IRIs, so that a huge number of queries can be generated easily.
There are many-to-one relations between CQ templates and SPARQL-OWL templates, since natural language allows us to define multiple synonymous forms of the same question. There are also  many-to-one relations between multiple query templates and a single CQ template, coming from cases in which there are multiple interpretations of the question i.e. \texttt{What is C1?} can be interpreted as a question about listing subclasses (providing definition by listing examples) or about listing superclasses (providing definition by putting the class in a broader context). %The dataset size obtained, allows us to answer positively to the \textbf{RQ1} -- we are able to generate a large-scale dataset of CQ template to SPARQL-OWL template pairs out of axiom shapes automatically.

\begin{figure}[h]
    \centering
    \includegraphics[width=0.85\textwidth]{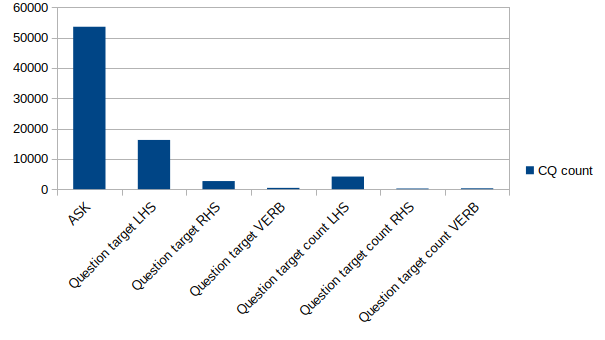}
    \caption{The number of CQ templates produced per query type}
    \label{fig:querytype}
\end{figure}

The dataset consists mainly, as presented in Figure~\ref{fig:querytype}, of ASK type CQs, since these we found the easiest to paraphrase in natural language. We found questions about numbers the hardest to paraphrase, so their representation is quite small (the category with the least number of questions, \textsc{Count query target RHS} contains 234 unique CQ templates). In Table~\ref{tab:constructs} we aggregated details on the OWL vocabulary used in SPARQL-OWL query templates in our dataset. We listed constructs that occured in more than 5\% of examples.

%To address research questions \textbf{RQ2} and \textbf{RQ3} 
We also analyzed how well \textsc{BigCQ} covers existing datasets. In order to do that, we checked if CQ templates from \textsc{BigCQ} match CQs from \textsc{CORAL}, that were not a part of \textsc{CQ2SPARQLOWL}. The results are presented in Table~\ref{tab:evaluation}. As we can see, there is a decent coverage of CQ forms. As one may suppose, the richness of expression possibilities of natural language makes it impossible to cover all forms of CQs. Moreover, some CQs stated in \textsc{CORAL} used words like: who, where, when, how often as question starters. These are not supported by our transformations, since the decision on which form should be used requires information on what is the type of the resource that fills the CQ template (e.g. if the type of a resource being a question target represents a person, the CQ should start with 'who' word).

The analysis of the coverage of SPARQL-OWL templates revealed that there is a high variety between the coverage depending on which from the 5 available ontologies in \textsc{CQ2SPARQLOWL} are checked. The highest coverage was obtained on queries from African Wildlife Ontology~\cite{DBLP:journals/corr/abs-1905-09519} (71\%), the lowest, on Stuff Ontology~\cite{stuff} (9\%). The analysis of errors showed that 19.38\% of all SPARQL-OWL queries from \textsc{CQ2SPARQLOWL} used the following query template:

\begin{verbatim}
    select ?x where {[] rdfs:subClassOf <C1>, [owl:onProperty ?x;
                                               owl:someValuesFrom [] ].} 
\end{verbatim}

which is not covered by our SPARQL-OWL templates. This kind of query however, was used only in the context of  Dem@Care~\cite{DBLP:conf/swat4ls/StavropoulosMKA15} ontology.

The next 11.63\% of the \textsc{CQ2SPARQLOWL} expected a query template that is also not covered by \textsc{BigCQ}:

\begin{verbatim}
    SELECT ?x WHERE {<C1> rdfs:subClassOf [a owl:Restriction ;
                                           owl:onProperty <OP1> ; 
                                           owl:someValuesFrom ?x ] .
                                          ?x rdfs:subClassOf <C2> }
\end{verbatim}

This kind of query was shared among multiple ontologies. The problem in that case comes from the fact that in \textsc{BigCQ} right-hand side of the property can be a variable, but there is no type restriction on that variable introduced. %but \textsc{BigCQ} does not add a restrict is no restriction on the type of that variable introduced. 
This functionality however, can be easily added in the next version of \textsc{BigCQ}. 
In the future, addressing only the described two cases can increase the coverage to over 75\%. The remaining 25\% of unsupported queries come from various situations, among which, we can distinguish:
\begin{enumerate}
    \item Using auxiliary variables that are not expected to be returned by the query.
    \item Using \texttt{union} instead of \texttt{owl:unionOf}.
    \item Using \texttt{owl:disjointWith} which is not observed among the most frequent axiom shapes.
    \item Asking about resources being a part of a complex class description (most often these are classes at the end of long property chains).
    \item Querying for more than one resource.
\end{enumerate}

\begin{table}
\caption{Overall summary of \textsc{BigCQ}}
\begin{center}
 \begin{tabular}{|l|r|} 
 \hline
 Measured dimension & measured value \\ 
 \hline
 Number of distinct CQ templates & 77575 \\ \hline
 Number of distinct SPARQL-OWL query templates & 549 \\ \hline
 Average number of CQ templates per SPARQL-OWL template & 171.68 \\ \hline
 Average number of SPARQL-OWL templates per a CQ & 1.22 \\ \hline
\end{tabular}
\label{tab:datasetsummary}
\end{center}
\end{table}

\begin{table}[]
\begin{center}
\caption{The number of the most occuring (present in more than 5\% of the datset) constructs among the list of all unique SPARQL-OWL query templates}
 \begin{tabular}{|r|r|} 
 \hline
 construct & times observed \\ 
 \hline
 \texttt{owl:Restriction} & 504/549 \\
 \texttt{owl:onProperty} & 504/549 \\
 \texttt{owl:intersectionOf} & 319/549 \\
 \texttt{rdfs:subClassOf} & 268/549 \\
 \texttt{owl:equivalentClass} & 281/549 \\
 \texttt{owl:someValuesFrom} & 273/549 \\
 \texttt{owl:qualifiedCardinality} & 87/549 \\
 \texttt{owl:hasValue} & 71/549 \\
 \texttt{owl:unionOf} & 69/549 \\
 \texttt{owl:allValuesFrom} & 56/549 \\
 \texttt{owl:maxQualifiedCardinality} & 38/549 \\
 \texttt{owl:complementOf} & 37/549 \\
 \texttt{owl:minCardinality} & 30/549 \\
 \hline
\end{tabular}
\end{center}
\label{tab:constructs}
\end{table}

\begin{table}
\begin{center}
\caption{The coverage of templates on real-world cases}
 \begin{tabular}{|r|r|} 
 \hline
 Dataset & Coverage \\ 
 \hline
 SPARQL-OWL queries from \textsc{CQ2SPARQLOWL} & 45.74\% \\ \hline
 CQs from \textsc{CORAL} & 63.89\% \\
 \hline
\end{tabular}
\end{center}
\label{tab:evaluation}
\end{table}

\section{The impact of \textsc{BigCQ}}
\label{impact}

%These, if filled with actual vocabulary from a given ontology can generate many times bigger dataset of materialized CQs. Even if we assume there is one way of filling the tempaltes, \textsc{BigCQ} provides more than 93x more CQs than the biggest to date \textsc{CORAL} requirements dataset. 

As we presented in Section\ref{sec:analysis}, \textsc{BigCQ} provides the biggest available dataset of CQ templates mapped to SPARQL-OWL templates. 

We focused on making the scripts generating \textsc{BigCQ} as easy to read and edit as possible. We hope that such approach will make it effortless to extend the number of axiom shapes, the number of transformations producing CQ templates from verbalizations and the database of synonymes that are all provided as simple textual representations. Thus, we hope that it will be interesting and easy for other researchers to improve the dataset, making it bigger and of better quality.

This dataset may be helpful for a broad range of research, like: providing training data for automatic CQ to SPARQL-OWL translators and Glossary of Terms extractors, improving controlled natural languages for CQ generation, automating ontology design methodologies and finally to understand the relation between CQs and ontologies better.

\section{Conclusions}
\label{sec:conclusions}
CQs are frequently used to gather requirements and to validate the quality of the ontology being constructed. The lack of large scale corpora of CQs with their formalizations in query languages make it hard for engineers to automate processes involving the use of CQs. Our dataset, \textsc{BigCQ} attempts to fill this gap by providing a large set of synthetically generated CQ templates matched with their SPARQL-OWL template forms. These templates may be used to be filled with resource labels and IRIs of actual resources from a given ontology to generate even bigger dataset, the size of which should be sufficient to fuel novel deep learning based approaches to information extraction or automatic translation from CQs into SPARQL-OWL. In this work, we decided to create our own solution to the problem of question generation from verbalizations. There are two reasons for this: (i) we wanted to obtain question templates that are constructed similarly to already collected sets of CQs. (ii) Existing question generation solutions focus on generating a limited number of questions. In our dataset, we wanted to obtain as big as possible dataset of CQ forms. (iii) Using the process described in the paper, it was easy to link CQ templates to SPARQL-OWL queries. We published~\footnote{\url{https://github.com/dwisniewski/BigCQ}} the dataset itself, the source code generating CQ to SPARQL-OWL template pairs from frequent axiom shapes as well as we provide a proof-of-concept code generating materialized CQ to SPARQL-OWL query pairs filled with labels and IRIs from a given ontology. We also generated the persistent URI to the dataset using Zenodo~\cite{dwisniewski_2021}.

\bibliographystyle{splncs04}
\bibliography{ms}

\begin{thebibliography}{10}
\providecommand{\url}[1]{\texttt{#1}}
\providecommand{\urlprefix}{URL }
\providecommand{\doi}[1]{https://doi.org/#1}

\bibitem{DBLP:conf/its/AldabeLMMU06}
Aldabe, I., et~al.: Arikiturri: An automatic question generator based on
  corpora and {NLP} techniques. In: Proc. of Intelligent Tutoring Systems,
  {ITS}. LNCS, vol.~4053, pp. 584--594. Springer (2006).
  \doi{10.1007/11774303\_58}

\bibitem{antoniou2004web}
Antoniou, G., Van~Harmelen, F.: Web ontology language: Owl. In: Handbook on
  ontologies, pp. 67--92. Springer (2004)

\bibitem{10.1007/978-3-030-21348-0_29}
Fern{\'a}ndez-Izquierdo, A., et~al.: {CORAL}: A corpus of ontological
  requirements annotated with lexico-syntactic patterns. In: Proc. of ESWC. pp.
  443--458. LNCS, Springer (2019)

\bibitem{FernndezLpez1997METHONTOLOGYFO}
Fern{\'a}ndez-L{\'o}pez, M., G{\'o}mez-P{\'e}rez, A., Juristo, N.:
  Methontology: From ontological art towards ontological engineering. In: AAAI
  1997 (1997)

\bibitem{DBLP:conf/rweb/FuchsKK08}
Fuchs, N.E., et~al.: Attempto controlled english for knowledge representation.
  In: Reasoning Web, International Summer School, Tutorial Lectures. LNCS,
  vol.~5224, pp. 104--124. Springer (2008). \doi{10.1007/978-3-540-85658-0\_3}

\bibitem{Gruninger1995MethodologyFT}
Gruninger, M.: Methodology for the design and evaluation of ontologies. In:
  IJCAI 1995 (1995)

\bibitem{Harris:13:SQL}
Harris, S., Seaborne, A.: {SPARQL} 1.1 query language. {W3C} recommendation,
  W3C (Mar 2013), http://www.w3.org/TR/2013/REC-sparql11-query-20130321/

\bibitem{kaljurand2007attempto}
Kaljurand, K.: Attempto controlled english as a semantic web language.
  University of Tartu (2007)

\bibitem{stuff}
Keet, C.: A core ontology of macroscopic stuff. In: Conference: International
  Conference on Knowledge Engineering and Knowledge Management. EKAW 2014.
  LNCS. pp. 209--224 (11 2014). \doi{10.1007/978-3-319-13704-9\_17}

\bibitem{DBLP:journals/corr/abs-1905-09519}
Keet, C.M.: The african wildlife ontology tutorial ontologies: requirements,
  design, and content. CoRR  \textbf{abs/1905.09519} (2019)

\bibitem{DBLP:conf/esws/KolliaGH11}
Kollia, I., et~al.: {SPARQL} query answering over {OWL} ontologies. In: Proc.
  of {ESWC}, Part {I}. pp. 382--396 (2011)

\bibitem{DBLP:journals/semweb/LawrynowiczPRT18}
Lawrynowicz, A., et~al.: Discovery of emerging design patterns in ontologies
  using tree mining. Semantic Web  \textbf{9}(4),  517--544 (2018).
  \doi{10.3233/SW-170280}

\bibitem{DBLP:conf/its/LiuCR10}
Liu, M., et~al.: Automatic question generation for literature review writing
  support. In: Intelligent Tutoring Systems, 10th International Conference,
  {ITS}, Proceedings, Part {I}. LNCS, vol.~6094, pp. 45--54. Springer (2010).
  \doi{10.1007/978-3-642-13388-6\_9}

\bibitem{DBLP:conf/semweb/MortensenHMN12}
Mortensen, J., et~al.: Modest use of ontology design patterns in a repository
  of biomedical ontologies. In: Proc. of the 3rd Workshop on Ontology Patterns,
  Boston, USA, November 12, 2012. {CEUR} Workshop Proceedings, vol.~929.
  CEUR-WS.org (2012)

\bibitem{DBLP:conf/esws/RenPMPDS14}
Ren, Y., et~al.: Towards competency question-driven ontology authoring. In:
  Proc. of The Semantic Web: Trends and Challenges, {ESWC}. LNCS, vol.~8465,
  pp. 752--767. Springer (2014). \doi{10.1007/978-3-319-07443-6\_50}

\bibitem{DBLP:conf/swat4ls/StavropoulosMKA15}
Stavropoulos, T.G., et~al.: {Dem@Care}: Ambient sensing and intelligent
  decision support for the care of dementia. In: Proc. of the Semantic Web
  Applications and Tools for Life Sciences International Conference. {CEUR}
  Workshop Proceedings, vol.~1546, pp. 229--230. CEUR-WS.org (2015)

\bibitem{Suarez-Figueroa2012}
Su{\'a}rez-Figueroa, M.C., et~al.: The NeOn Methodology for Ontology
  Engineering, pp. 9--34. Springer Berlin Heidelberg, Berlin, Heidelberg
  (2012). \doi{10.1007/978-3-642-24794-1\_2}

\bibitem{DBLP:conf/semweb/WangPH06}
Wang, T.D., et~al.: A survey of the web ontology landscape. In: Proc. of The
  Semantic Web - {ISWC} 2006, 5th International Semantic Web Conference, {ISWC}
  2006. LNCS, vol.~4273, pp. 682--694. Springer (2006).
  \doi{10.1007/11926078\_49}

\bibitem{DBLP:conf/lats/WangLNWGB18}
Wang, Z., et~al.: {QG-net}: a data-driven question generation model for
  educational content. In: Proc. of the Fifth Annual {ACM} Conference on
  Learning at Scale. pp. 7:1--7:10. {ACM} (2018). \doi{10.1145/3231644.3231654}

\bibitem{10.1145/3184558.3186575}
Wisniewski, D.: Automatic translation of competency questions into {SPARQL-OWL}
  queries. In: Proc. of WWW, Companion volume. p. 855–859 (2018)

\bibitem{dwisniewski_2021}
Wisniewski, D.: {BigCQ}  (Apr 2021). \doi{10.5281/zenodo.4704674}

\bibitem{10.1007/978-3-030-32327-1_36}
Wisniewski, D., {\L}awrynowicz, A.: A tagger for glossary of terms extraction
  from ontology competency questions. In: Proc. of {ESWC}, Satellite Events.
  pp. 181--185. Springer (2019)

\bibitem{DBLP:journals/ws/WisniewskiPLK19}
Wisniewski, D., et~al.: Analysis of ontology competency questions and their
  formalizations in {SPARQL-OWL}. JWS  \textbf{59} (2019)

\end{thebibliography}

\end{document}